\documentclass{article}

    \PassOptionsToPackage{numbers, compress}{natbib}
\usepackage[preprint]{neurips_2019}

\usepackage[utf8]{inputenc} % allow utf-8 input
\usepackage[T1]{fontenc}    % use 8-bit T1 fonts
\usepackage{hyperref}       % hyperlinks
\usepackage{url}            % simple URL typesetting
\usepackage{booktabs}       % professional-quality tables
\usepackage{amsfonts}       % blackboard math symbols
\usepackage{nicefrac}       % compact symbols for 1/2, etc.
\usepackage{microtype}      % microtypography

\usepackage{microtype}
\usepackage{graphicx}
\usepackage{subcaption}
\usepackage{booktabs} % for professional tables
\usepackage{wrapfig}

\usepackage{floatrow}
\newfloatcommand{capbtabbox}{table}[][\FBwidth]

\usepackage{amsfonts}
\usepackage{amssymb}
\usepackage{amsmath}
\usepackage{multirow}

\makeatletter
\newcommand{\ssymbol}[1]{$^{\@fnsymbol{#1}}$}
\makeatother

\title{Unsupervised Pre-training for Natural Language Generation: A Literature Review}

\author{%
  Yuanxin Liu\textsuperscript{1,2},
  Zheng Lin\textsuperscript{1}\\
  \textsuperscript{1}Institute of Information Engineering, Chinese Academy of Sciences\\
  \textsuperscript{2}School of Cyber Security, University of Chinese Academy of Sciences\\
}

\begin{document}

\maketitle

\begin{abstract}
Recently, unsupervised pre-training is gaining increasing popularity in the realm of computational linguistics, thanks to its surprising success in advancing natural language understanding (NLU) and the potential to effectively exploit large-scale unlabelled corpus. However, regardless of the success in NLU, the power of unsupervised pre-training is only partially excavated when it comes to natural language generation (NLG). The major obstacle stems from an idiosyncratic nature of NLG: Texts are usually generated based on certain context, which may vary with the target applications. As a result, it is intractable to design a universal architecture for pre-training as in NLU scenarios. Moreover, retaining the knowledge learned from pre-training when learning on the target task is also a non-trivial problem.
% As a result, it is impossible to design a universal context representation extractor, and the pre-trained text generators might have to modify their inner structures to incorporate specific context information, which could have remained mostly untouched in NLU scenarios.
% As a result, when being applied to a downstream task of conditional NLG, the pre-trained models have to cater for specific input context(s) by modifying their inner structures, which could have remained mostly untouched in NLU scenarios. 
% This review summarizes the recent efforts to enhance NLG systems with unsupervised pre-training, which, as we observed, are means to catalyse the integration of pre-trained models into downstream tasks. We propose to classify the studies into architecture-based methods and strategy-based methods.
This review summarizes the recent efforts to enhance NLG systems with unsupervised pre-training, with a special focus on the methods to catalyse the integration of pre-trained models into downstream tasks. They are classified into architecture-based methods and strategy-based methods, based on their way of handling the above obstacle. Discussions are also provided to give further insights into the relationship between these two lines of work, some informative empirical phenomenons, as well as some possible directions where future work can be devoted to.

% the characteristics of texts generated by models with pre-training, the implications of unsupervised pre-training for NLG,

\end{abstract}

\section{Introduction}
\label{sec:intro}
Unsupervised pre-training has sparked a sensational research interest in the natural language processing (NLP) community. This technology provides a promising way to exploit linguistic information from large-scale unlabelled textual data, which can serve as an auxiliary prior knowledge to benefit a wide range of NLP applications. In the literature, language modeling (LM) is a prevalent task for pre-training, where the target words are predicted conditioned on a given context.
% Typically, pre-training is conducted in a generative fashion, where target words are predicted conditioned on a given context, e.g., language model (LM). 
Therefore, it is intuitive to employ the pre-trained LMs for natural language generation, as the pre-training objective naturally accords with the goal of NLG. However, revolutionary improvements are only observed in the field of NLU. 

\par The primary factor that impedes the progress of unsupervised pre-training in NLG is an idiosyncratic nature of text generation: Basically, we do not write words from scratch, but instead based on particular context, e.g., the source language sentences for translation, the dialog histories for response generation, and the visual scenes for image captioning, among others. In unsupervised pre-training, the task-specific context is not available, which leads to a discrepancy between pre-training and training in the target task. More precisely, the challenges posed by the discrepancy can be reflected in two aspects: First, the diverse context makes it intractable to design a universal representation extractor as in the case of NLU, and the pre-trained language generators may have to modify their inner structures to deal with the task-specific context. Second, the mismatch in data distribution and objective between the two training stages might result in the performance on the pre-training tasks being compromised during fine-tuning, which is dubbed as the \textit{catastrophic forgetting} problem \cite{Goodfellow13}.
% It is difficult to consider the diverse context in pre-training strategy, which, inevitably leads to a discrepancy between unsupervised pre-training and supervised training in the target task.

\par In response to the above challenges, two lines of work are proposed by resorting to architecture-based and strategy-based solutions, respectively. 
% To facilitate the consistency between the two training stages, some studies focused on designing appropriate model architectures, either for task-specific purposes or task-agnostic purposes. 
Architecture-based methods either try to induce task-specific architecture during pre-training (task-specific methods), or aim at building a general pre-training architecture to fit all downstream tasks (task-agnostic methods). 
% can be further divided into for task-specific purposes and for task-agnostic purposes. The former tries to induce task-specific architecture during pre-training, while the latter aims at building a general pre-training architecture to fit all downstream tasks. 
Strategy-based methods depart from the pre-training stage, seeking to take advantage of the pre-trained models during the process of target task learning. The approaches include fine-tuning schedules that elaborately design the control of learning rates for optimization, proxy tasks that leverage labeled data to help the pre-trained model better fit the target data distribution, and knowledge distillation approaches that ditch the paradigm of initialization with pre-trained parameters by adopting the pre-trained model as a teacher network.
% resort to elaborately designed fine-tuning schemes to bridge the discrepancy. 
% proxy tasks that help the pre-trained model to better fit the distribution of labelled data, decoding schemes that cater for the pre-training objectives like masked language model (it will be described in Section \ref{sec:background}), and knowledge distillation approaches that bypass the difficulty of direct transfer by adopting the pre-trained model as an assistant module.

\par The remainder of this review is organized as follows: In Section \ref{sec:background}, we will introduce the background knowledge about unsupervised pre-training for NLU, followed by a sketch of how the pre-trained models are employed through parameter initialization for NLG in Section \ref{sec:direct}. In Section \ref{sec:architecture}, we will describe the architecture-based methods, and the strategy-based methods are presented in Section \ref{sec:strategy}. Section \ref{sec:discussion} provides some in-depth discussions, and Section \ref{sec:conclusion} concludes this review.

%%%%%%%%%%%%%%%%%%%%%%%%%%%%%%%%%%%%%%%%%%%%%%%
\section{Background: Unsupervised Pre-training for NLU}
\label{sec:background}
Learning fine-grained language representations is a perennial topic in natural language understanding. In restrospect, compelling evidences suggest that good representations can be learned through unsupervised pre-training. 

% \par Early work focused on word-level representations \cite{MikolovSCCD13,PenningtonSM14}, which encodes each word independently. For sentence-level representations, there are roughly two kinds of pre-training objectives, namely discriminative pre-training and generative pre-training. Discriminative pre-training tries to capture the semantic relationship between sentences, usually by performing a binary classification task: distinguishing context sentence(s) for a given sentence from non-context sentence(s) \cite{DevlinCLT19,LogeswaranL18}. Generative pre-training follows the language model paradigm:

\par Early work focused on word-level representations \cite{MikolovSCCD13,PenningtonSM14}, which encodes each word independently. For sentence-level representations, there are roughly two kinds of pre-training objectives, namely discriminative pre-training and generative pre-training. Discriminative pre-training distinguishes context sentence(s) for a given sentence from non-context sentence(s) \cite{DevlinCLT19,LogeswaranL18}, with the aim to capture inter-sentence relationships. Generative pre-training follows the language model paradigm:
\begin{equation}
\max_{\theta} \sum_{t=1}^{T} \log P\left(x_{t} | C; \theta\right)
\end{equation}
where $x_{t}$ is the $t^{th}$ word in the textual sequence to generate, $T$ indicates sequence length, $\theta$ stands for learnable parameters, and $C$ is the context information, which is defined by the pre-training objective. ELMo \cite{PetersNIGCLZ18} and GPT (short for Generative Pre-training) \cite{Radford18} adopt uni-directional Transformer \cite{VaswaniSPUJGKP17} and bi-directional LSTM \cite{HochreiterS97} language models, respectively. In this case, the context is defined as $x_{1:t}$ or $x_{t+1:T}$. BERT \cite{DevlinCLT19} is trained with a novel masked language model (MLM), which is a non-autoregressive way of generation. Specifically, MLM randomly replaces a fixed proportion of tokens in each sentence with a special [MASK] token or a random token, which results in a corrupted sentence $X_{\text{mask}}$, and predicts each replaced token based on the same context $X_{\text{mask}}$. To alleviate the inconsistency with target tasks caused by the introduction of [MASK] token, XLNet \cite{Yangzhilin} introduces permutation-based language model, which conducts autoregressive language modeling over all possible permutations of the original word sequence. This gives rise to a context $C=X_{\mathbf{z}_{1:t-1}}$, where $\mathbf{z}$ is a certain permutation of $[1,2, \ldots, T]$, according to the definitions in \cite{Yangzhilin}. \cite{DaiL15} and \cite{KirosZSZUTF15} pre-trained an encoder-decoder framework to reconstruct the input sentence and the surrounding sentence, respectively, and the encoded input sentence thereby is included in the context $C$.
% To achieve sentence-level representation, \cite{DaiL15} and \cite{KirosZSZUTF15} proposed pre-training tasks that use the encoder-decoder framework to reconstruct the input sentence and the surrounding sentences, respectively. \cite{PetersNIGCLZ18} took into account contextual information to model word representations via bi-directional LMs, and revealed that hidden states at different levels of the pre-trained LMs convey different aspects of semantic. The most recent advances followed the ``pre-training and fine-tuning" paradigm \cite{Radford,DevlinCLT19,Yang}, where the pre-trained models will be fine-tuned jointly with the supervised task.

\par The sentence representations learned by LMs \footnote{In the rest of this review, we refer to unsupervised pre-training as the generative LMs, unless otherwise specified.} can be used to perform many NLU tasks by adding a simple linear classifier. Despite the objective of language modeling, the pre-trained representations and have successfuly pushed the state-of-the-art on multiple benchmarks . 
% One possible explanation is that the pre-trained model can be easily integrated into the intended task via several task-specific linear transformation matrices, with its inner structure mostly untouched.

% \section{Direct Employment of Pre-trained LMs in NLG}
% \section{Employing Pre-trained LMs via Initialization for NLG}
\section{Unsupervised Pre-training and Parameter Initialization for NLG}
\label{sec:direct}
NLG systems are usually built with an encoder-decoder framework, where the encoder reads the context information and the decoder generates the target text from the encoded vectorial representations. A direct way to utilize the pre-trained models is to initialize part of the encoder (when dealing with textual context) and/or the decoder with pre-trained parameters.
% \footnote{Generative pre-training plays a dominant role in current studies on unsupervised pre-training for NLG, therefore we use LMs in our discussions for simplicity.}. 
For the encoder, pre-training is expected to provide better sentence representations, as we discussed in Section \ref{sec:background}. For the decoder, the intuition is to endow the model with some rudimentary ability for text generation.

\par \cite{LiuYang} employed BERT as the encoder for abstractive text summarization, with some additional techniques to help integrate the BERT-initialized encoder with the randomly initialized decoder, which we will explicate in Section \ref{fine-tune}. GPT-2 \cite{Radford19} inherited the left-to-right LM pre-training objective from GPT and extended the application to NLG, where the pre-trained LM directly serves as the language generator, with some special symbols to identify task-specific contexts. In the case of zero-shot task transfer, preliminary experiments showed that straightforward adaption of GPT-2 compares unfavorably with other unsupervised baselines. 

\par \cite{RamachandranLL17} is among the first attempts to investigate unsupervised pre-training for sequence to sequence (Seq2Seq) learning. They used pre-trained LSTM-based LMs to initialize the first layer of the encoder and the decoder, which act as representation extractors. An additional LSTM layer, which is randomly initialized, is then added on top of the pre-trained LMs to build the Seq2Seq framework. To make use of the text generation ability of LMs, the output softmax layer of the decoder LM is also retained. 
% The word embeddings and the output softmax layer of the decoder are also pre-trained. 
% \cite{RamachandranLL17} also proposed some architecture-related refinements and showed that jointly optimizing LM and the Seq2Seq objective is beneficial. As our main focus is unsupervised pre-training, in this review we will not probe into the technical details of each work. 
Some recent endeavours \cite{Lample19,Rothe191} explored multiple combinations of GPT- and BERT-based models to initialize the encoder and the decoder, respectively. Although remarkable results are observed, the separately pre-trained LMs are still inconsistent with the Seq2Seq framework.
% Taking machine translation as an example, they pre-train two LSTM-based LMs on the source language corpus and the target language corpus, respectively. An additional LSTM layer, which is randomly initialized, is then added on top of the pre-trained LMs to build the Seq2Seq framework. The word embeddings and the output softmax layer of the decoder are also pre-trained. For monolingual applications, one pre-trained LM is used to initialize both the encoder and the decoder. \cite{RamachandranLL17} also proposed some architecture-related refinements and showed that jointly optimizing LM and the Seq2Seq objective is beneficial. As our main focus is unsupervised pre-training, in this review we will not probe into the technical details of each work. 

\section{Architecture-based Methods}
\label{sec:architecture}
\subsection{Inducing Task-Specific Architecture in Pre-training}
Separately initializing the encoder and the decoder with LMs neglects the interaction between the two modules at the pre-training stage, which is sub-optimal. For NLG tasks that can be modeled as Seq2Seq learning, it is feasible to jointly pre-train the encoder and the decoder. Existing approaches for this sake can be categorized into three variants: Denoising autoencoders (DAEs), conditional masked language models (CMLMs) and sequence to sequence language models (Seq2Seq LMs).

%%%%%%%%%%%%%%%%%%%%%%%%%%%%%%%%%%%%%%%%%%%%%%
\begin{figure*}[t]
\centering
\includegraphics[width=1.0\textwidth]{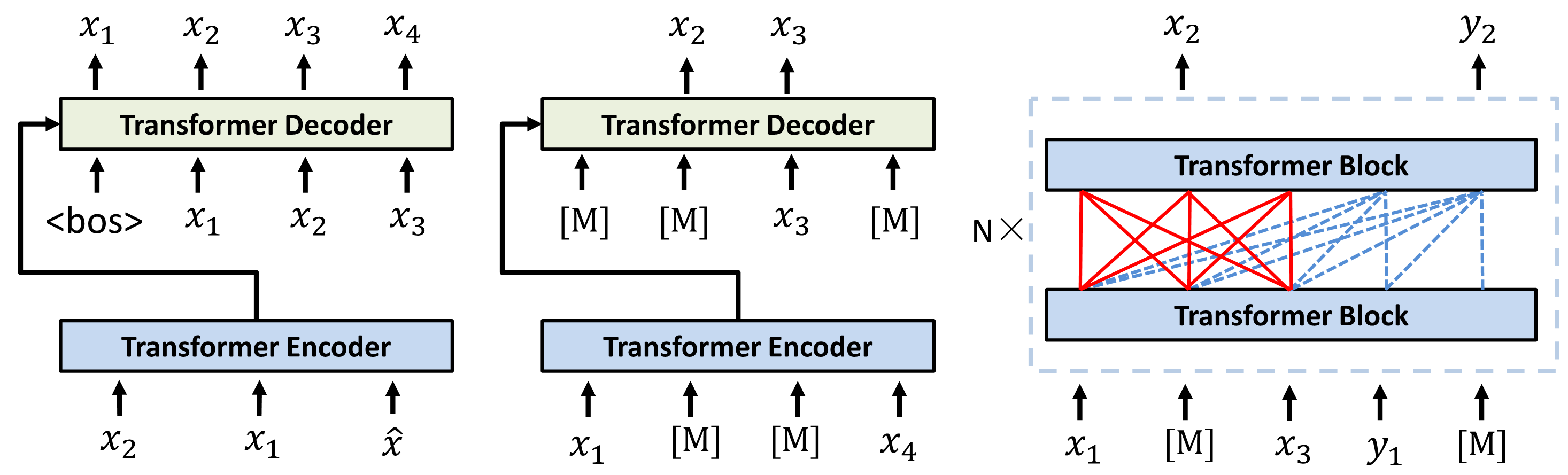} 
\caption{Overview of pre-training tasks for Seq2Seq learning. \textbf{Left:} Denoising autoencoder \cite{Wang} takes a corrupted sentence as input and reconstructs the original sentence. The positions of $x_{1}$ and $x_{2}$ are switched by \textit{shuffling}, $x_{3}$ is \textit{replaced} with the token $\hat{x}$, and $x_{4}$ is \textit{deleted}. \textbf{Middle:} Conditional masked language model \cite{SongTQLL19} masks several consecutive tokens in a sentence before feeding it to the encoder, and the input sentence to the decoder is constructed with the unmasked tokens in the encoder side being masked. \textbf{Right:} Seq2Seq language model \cite{Dong} is composed of a single Transformer model, which takes the concatenation of a source sentence and a target sentence as input. The special tokens inserted between sentences are omitted for conciseness. The tokens in the source sentence can attend to each other, while each token in the target sentence can only attend to the source tokens and the target tokens on its left. The MLM-like training objective is used.}
\label{fig:structure}
\end{figure*}
%%%%%%%%%%%%%%%%%%%%%%%%%%%%%%%%%%%%%%%%%%%%%%

\subsubsection{Denoising Autoencoder} An intuitive way to conduct unsupervised Seq2Seq learning is to train an autoencoder (AE) based on encoder-decoder framework. Different from AEs, DAEs take a corrupted sentence as input and reconstruct the original sentence. The advantage is that the corrupted input will force the decoder to extract relevant information from the source side for text generation. To obtain the corrupted sentence, \cite{Wang} designed three noising functions: \textit{shuffle}, \textit{delete} and \textit{replace} (the left plot of Figure \ref{fig:structure} gives an illustration), each of which is controlled by a pre-defined probability distribution. To be more specific, each token in the raw sequence is assigned with a new index based on a gaussion distribution $N(0, \sigma)$; the delete and replace operations of a token are determined by a Bernoulli distribution $B(p)$ with Beta distribution as prior. The three functions are applied to the raw sequences in random order.

\subsubsection{Conditional Masked Language Model} CMLM \cite{SongTQLL19} extends the single model MLM proposed by \cite{DevlinCLT19} to the encoder-decoder setting, where the masked text sequence is read by the encoder, and the decoder only reconstructs the masked tokens, in construct to the entire sequence in DAEs. As the middle plot of Figure \ref{fig:structure} shows, CMLM masks consecutive tokens \footnote{CMLM can also be implemented in other manners \cite{Anonymous,Marjan}, where discrete tokens in the textual sequence are masked.}, and the unmasked tokens in the encoder side are masked when being feed to the decoder. Following the notations in \cite{SongTQLL19}, let us assume that the tokens with index from $u$ to $v$ are masked from the raw sentence $X$, which results in $X^{\backslash u: v}$, and $X^{u: v}$ denotes the decoder input. Then, when predicting each masked token $x_{t}$ ($u \leq t \leq v$), the context is $X^{u: v}_{<t}$ and $X^{\backslash u: v}$. The underlying motivation, as \cite{SongTQLL19} argued, is to force the encoder to understand the meaning of the unmasked tokens, which is achieved by encoder side masks, and encourage the decoder to refer to the source information rather than the leftward target tokens, which is achieved by decoder side masks.

\subsubsection{Sequence to Sequence Language Model} Seq2Seq LM \cite{Dong} performs Seq2Seq modeling using a single Transformer model, with the concatenation of source sentence and target sentence as input. To simulate Seq2Seq learning with encoder-decoder frameworks, the attention span of each target token is constrained to the source tokens and the leftward target tokens, which is achieved by self-attention masks (see the right plot of Figure \ref{fig:structure}). In this way, the ability to extract language representation and generate texts are integrated into a single model. It is worth mentioning that Seq2Seq LM does not auto-regressively generate the target sentence, but instead predicting masked tokens based on the contexts controlled by self-attention masks. In other words, Seq2Seq LM still belongs into the family of MLMs. Apart from Seq2Seq LM, \cite{Dong} also explored uni-directional LM and bi-directional LM structures to perform the MLM-based cloze task, and incorporated the three kinds of LMs to build the final pre-training objective.

\subsection{Encoder-Agnostic Architectures for Adaptation}
\begin{figure*}[t]
\centering
\includegraphics[width=1.0\textwidth]{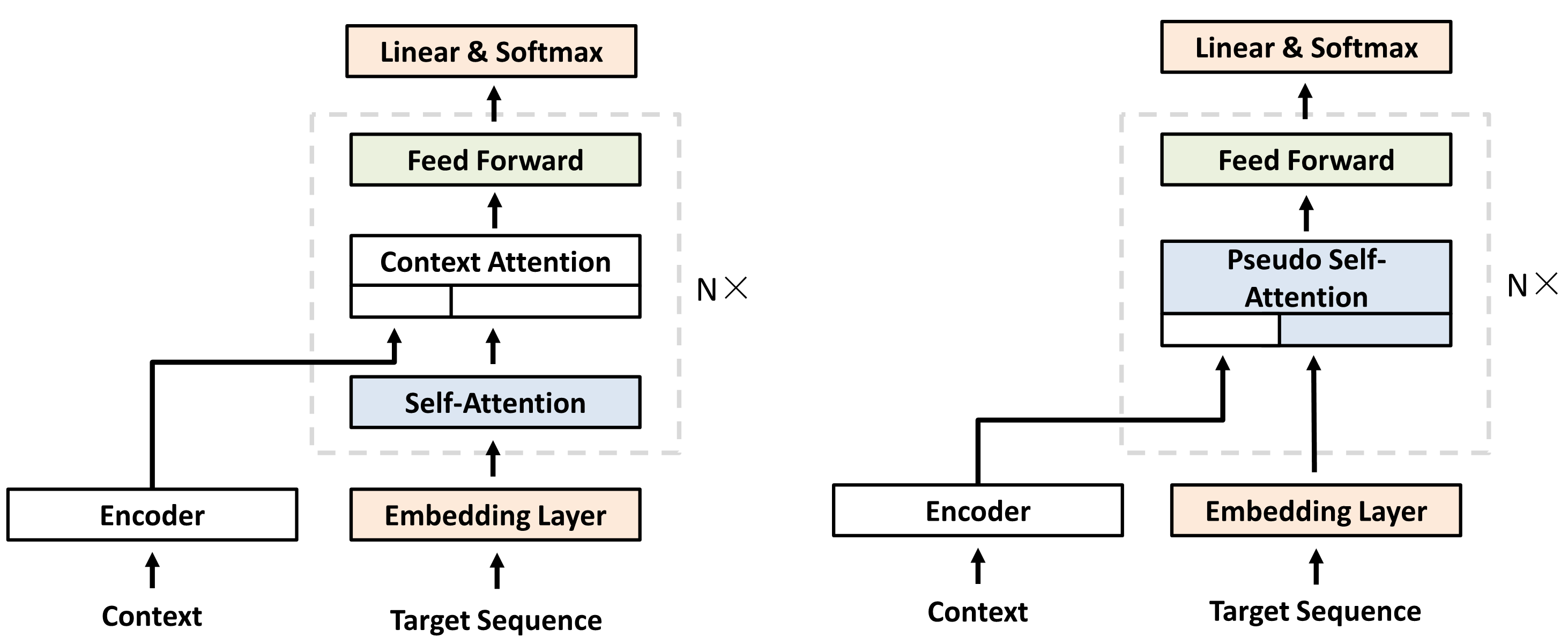} 
\caption{Overview of two encoder-agnostic architecture variants. The solid blocks with white backgrounds are randomly initialized. Residual connection and layer normalization are omitted for simplicity.  \textbf{Left:} Context Attention follows the Transformer decoder side attention over the encoder outputs. \textbf{Right:} The encoder outputs are incorporated as part of the decoder side Pseudo Self-Attention.}
\label{fig:agnostic}
\end{figure*}

Although the Seq2Seq-based pre-training methods exhibit strong performance, they are confined to text-to-text generation. In order to encompass more diverse contexts, some researches began to investigate encoder-agnostic pre-training architectures \cite{Golovanov19,Ziegler19}. \textit{Context Attention} and \textit{Pseudo Self-Attention} are two typical variants presented by \cite{Ziegler19}, which differ in the way that the task-specific context is injected (see Figure \ref{fig:agnostic}). Context Attention takes the form of a standard Transformer decoder, with the layer that attends to the encoder outputs being randomly initialized. Pseudo Self-Attention considers the context vectors and the previous layer decoder outputs as an integral input, and the attended results are computed as follows:
\begin{equation}
\operatorname{PSA}(C, Y)=\operatorname{softmax}\left(\left(Y W_{q}\right)\left[\begin{array}{c}{C W^{c}_{k}} \\ {Y W^{y}_{k}}\end{array}\right]^{\top}\right)\left[\begin{array}{c}{C W^{c}_{v}} \\ {Y W^{y}_{v}}\end{array}\right]
\end{equation}
where $C \in \mathbb{R}^{|C| \times d_{c}}$ and $Y \in \mathbb{R}^{|Y| \times d_{y}}$ are the context vectors and representations of the target textual sequence, respectively. The linear transformation matrices $W^{c}_{k}, W^{c}_{v} \in \mathbb{R}^{|C| \times d_{model}}$ with respect to $C$ are added to project the context to the self-attention space, and $W_{q}, W^{y}_{k}, W^{y}_{v} \in \mathbb{R}^{|Y| \times d_{model}}$ are part of the pre-trained model.
\par Except for the performance on target tasks, an alternative metric to gauge the quality of encoder-agnostic architectures is the degree to which the pre-trained parameters have to change, in order to inject the task-specific context. \cite{Ziegler19} compared the parameter changes of Context Attention and Pseudo Self-Attention in the feed forward layer, and discovered that Pseudo Self-Attention is more robust under this evaluation.

%%%%%%%%%%%%%%%%%%%%%%%%%%%%%%%%%%%%%%%%%%%%%%
%%%%%%%%%%%%%%%%%%%%%%%%%%%%%%%%%%%%%%%%%%%%%%

\section{Strategy-based Methods}
\label{sec:strategy}

\subsection{Fine-tuning Schedules for Adaption}
\label{fine-tune}
When the pre-trained model is only a part of the target task system, fine-tuning requires joint learning of the components initialized in different fashion, which can make the training process unstable. The pre-trained model may also suffer from aggravated catastrophic forgetting problem as it has to coordinate with other components during fine-tuning \cite{EdunovBA19,Yangjiacheng}.  
% Most recent unsupervised pre-training algorithms followed the ``Pre-training and fine-tuning'' procedure. In NLU tasks, fine-tuning based methods outperform fixed representations \cite{DevlinCLT19}. However, preliminary experiments in \cite{EdunovBA19} show that the pre-trained LM representations yield better results without fine-tuning. This is conceivable as NLG fine-tuning requires coordination of the encoder and decoder, which can be trained and/or initialized in different fashion. The interdependence between the two modules with different start-points makes fine-tuning unstable and may exacerbate the catastrophic forgetting problem. 
% Most recent unsupervised pre-training algorithms followed the ``Pre-training and fine-tuning'' procedure. However, the mismatch in data distribution and optimization objective between pre-training and fine-tuning could make training unstable. For example, when \cite{RamachandranLL17} fine-tuned the separately pre-trained LMs to Seq2Seq learning tasks, they observed that the performance on language modeling severely deteriorated. This phenomenon is referred to as \textbf{catastrophic forgetting}, which could also compromise the performance on downstream tasks. 
From the perspective of optimization, it is unreasonable to schedule the pre-trained components and the newly-introduced components with the same learning rate, considering that the former have already possessed some unique knowledge. A common assumption is that the pre-trained parameters should be updated at a slower learning rate and with smoother decay \cite{LiuYang,Yangjiacheng}. The rationale behind such setting is that fine-tuning with more accurate gradient can prevent the pre-trained parameters from deviating too faraway from the original point, and the newly-introduced components need to quickly converge to the target parameter space. 
% A common assumption is that the randomly initialized parameters should be updated at a faster learning rate to convergence, and the pre-trained parameters should be trained with more accurate gradient so as not to deviate too faraway from original region \cite{LiuYang,Yangjiacheng}. 
To this end, \cite{LiuYang} adopted two Adam optimizers with different learning rates for the pre-trained encoder and the randomly initialized decoder. The learning rates are scheduled as in \cite{VaswaniSPUJGKP17} with different warming up steps:
\begin{equation}
\begin{array}{l}{l r_{\operatorname{Enc}}=\tilde{l} r_{\operatorname{Enc}} \cdot \min \left(\operatorname{step}^{-0.5}, { step } \cdot { warmup }_{\operatorname{Enc}}^{-1.5}\right)} \\ {l r_{\operatorname{Dec}}=\tilde{l} r_{\operatorname{Dec}} \cdot \min \left({step}^{-0.5},  { step } \cdot  { warmup }_{\operatorname{Dec}}^{-1.5}\right)}\end{array}
\end{equation}
where ${warmup}_{\operatorname{Enc/Dec}}$ and $\tilde{l}r_{\operatorname{Enc/Dec}}$ determine the speed of learning rate changes and the max learning rates, respectively.

% \subsection{Decoding Schemes for Adaption}

\subsection{Proxy Tasks for Adaption}
Large-scale unlabelled data provides generic linguistic knowledge, but the target tasks have unique data distribution and objectives. An effective way to bridge this gap is to introduce proxy tasks with moderate changes to the pre-training objectives, but at the same time take the labeled data into account \cite{Lample19,Anonymous}. \textit{Translation Language Modeling} (TLM) \cite{Lample19} is a special generalization of MLM in the cross-lingual situation. It leverages the paralleled machine translation corpus for further training of the LMs that are pre-trained on monolingual corpora. Specifically, the source language sentence and the corresponding target language sentence are fed to the model in parallel, with random tokens from each language being masked to perform the cloze-style prediction as in MLM. Different from monolingual MLM, TLM encourages word predictions to rely on the interdependence from two languages, therefore the sentence representations learned from separate languages can be well aligned. 

\par For some particular NLG tasks, existing proxy tasks designed under the supervised setup can also work with unsupervised pre-training models. For instance, in neural text summarization, the combination of extractive and abstractive\footnote{Extractive text summarization selects sub-sequences from the source texts, while abstractive text summarization treats the task as a Seq2Seq learning problem.} objectives can generate better summaries \cite{LiXLW18,GehrmannDR18}. 
% In text summarization, extractive summarization and abstractive summrization are two dominant forms of approaches. The former selects sub-sequence from the source texts, while the later treats the task as a Seq2Seq learning problem. It has been shown that combining the extractive objective and abstractive objective can generate better summaries \cite{LiXLW18}. 
Inspired by this, \cite{LiuYang} introduced extractive summarization as a proxy task to fine-tune the pre-trained BERT, before adopting it as the abstractive summarization encoder. Compared with the original BERT features, the representations learned from extractive summarization contain more task-specific information, therefore conveying the meaning of source texts better.

\subsection{Knowledge Distillation for Adaption}
The aforementioned methods are diverse in  implementation, but share the common idea of employing the pre-trained models through parameter initialization. 
% As we discussed above, the fine-tuning process may lead to some optimization difficulties and the pre-trained model with large size are not applicable to systems that have limited computational resources. 
An alternative way to exploit the pre-trained models is using the knowledge distillation technique \cite{HintonVD15}. Knowledge distillation is a special form of training, where a \textit{student} network learns from the supervision signals produced by a \textit{teacher} network. 
% Specifically, the supervision signals can be defined as the probability distribution predicted by the teacher network \cite{HintonVD15,Anonymous}, or directly derived from the hidden representations \cite{Yangjiacheng}. 
\par Taking BERT as an example, the pre-trained MLM contains global information, which can teach the autoregressive Seq2Seq models to ``see from the future'' \cite{Anonymous}. In practice, the probability distribution predicted by BERT is regarded as a soft label to compute the cross-entropy loss function :
% The teacher probability distribution can be seen as a soft label, as compared with the hard one-hot label, to compute the loss function:
%
\begin{equation}
\mathcal{L}^{cross}_{kd}(\theta)=-\sum_{t=1}^{|Y|}\sum_{w \in \mathcal{V}}\left[P(y_{t}=w | Y^{masked}, X ; \phi ) \cdot \log P\left(y_{t}=w | Y_{1: t-1}, X ; \theta \right)\right]
\end{equation}
where $X$, $Y$ and $Y^{masked}$ are the source sequence, the raw target sequence and the masked target sequence, respectively. $\mathcal{V}$ denotes the output vocabulary. $\theta$ indicates the parameters of the student network (Seq2Seq), which are learnable, and $\phi$ indicates the BERT parameters, which are fixed. In this way, the knowledge from unsupervised pre-training can be flexibly transferred to the target tasks, dispensing with the size and architecture limitations. 

\par The supervision can also be derived from the hidden representations \cite{Yangjiacheng}, with a mean-squared-error (MSE) distillation loss:
\begin{equation}
\mathcal{L}^{mse}_{kd}=-\left\|{h}^{bert}_{m}-h^{seq2seq}_{n}\right\|_{2}^{2}
\end{equation}
where $m$ and $n$ are hyper-parameters denoting the layer subscripts. Compared with the probability soft labels, the representation distillation method requires the Seq2Seq model to have the same hidden size with BERT, which is a more strict constrain.

\par Combining the knowledge distillation loss and the standard generative loss for Seq2Seq learning gives rise to the final objective to optimize:
\begin{equation}
\mathcal{L}(\theta)=\alpha \mathcal{L}_{kd}(\theta)+(1-\alpha) \mathcal{L}_{seq2seq}(\theta)
\end{equation}
where $\alpha$ is the weighting term that balances the contribution of the two kinds of loss functions.

\section{Discussions}
\label{sec:discussion}
\subsection{The Relationship between Architecture- and Strategy-based Methods}
We have analysed two major challenges faced by the application of unsupervised pre-training to NLG (see Section \ref{sec:intro}). On this basis, we introduced existing methodologies from the architecture and strategy considerations. The architecture-based methods are mainly proposed in response to the first challenge. Since the architecture of pre-trained model has a significant effect on the downstream task (when the pre-trained parameters are used for initialization), architecture designings have to plan in advance to narrow the discrepancy between pre-training and training on target tasks. This motivation has shown great effectiveness on the Seq2Seq framework \cite{Wang,SongTQLL19,Dong}. The strategy-based methods focus on the second challenge. They take a postprocessing point of view, with the aim to make the best of the pre-trained model at the target task training stage. It is noteworthy that the challenges are not independent inherently, and the two types of methods can actually work as complement to each other. For example, the fine-tuning schedules can alleviate the negative effects caused by the modification of pre-trained structures, and the catastrophic forgetting problem can also seek solution by devising a general task-agnostic architecture.

\subsection{Experimental Phenomenons}
Existing researches on unsupervised pre-training for NLG are conducted on various tasks for different purposes. Probing into the assorted empirical results may help us discover some interesting phenomenons:
\begin{itemize}
\item The advantage of pre-training gradually diminishes with the increase of labeled data \cite{RamachandranLL17,Wang,SongTQLL19}. 

\item Fixed representations yield better results than fine-tuning in some cases \cite{EdunovBA19}.

\item Overall, pre-training the Seq2Seq encoder outperforms pre-training the decoder \cite{EdunovBA19,Wang,Lample19,Rothe191}. 
\end{itemize}

\par The first two phenomenons attest to the catastrophic forgetting theory. Thanks to the access to large-scale unlabeled corpora, unsupervised pre-training is able to excel  at zero/low-shot settings, while the pre-trained models can only achieve few gains when abundant labeled data is available. This can be explained by the high quality of the dataset and the capacity of the task-specific models, which leave little space for improvement. Nonetheless, the increased supervision from labeled data can also influence the performance on pre-training tasks. By fixing the pre-trained parameters, the learned representations will not be affected by the numerous iterations of training on the target task, which makes them work better without fine-tuning.

\par The third phenomenon is kind of counter-intuitive, as the generative pre-training objectives are more similar to the decoder's function. There is no unanimous theory to explain why the encoder is a more important element to pre-train. But this discovery suggests that the pre-trained LMs are more robust when acting as representation extractors, while they are more sensitive the the change of context when acting as conditional language generators.

% \subsection{In what way does pre-training make the difference?}
% \subsection{The Implications of Unsupervised Pre-training for NLG}
\subsection{Future Directions}
The diversity of NLG applications poses challenges on the employment of unsupervised pre-training, yet it also raises more scientific questions for us to explore. In terms of the future development of this technology, we emphasize the importance of answering four questions: 1) How to introduce unsupervised pre-training into NLG tasks with cross-modal context? 2) How to design a generic pre-training algorithm to fit a wide range of NLG tasks? 3) How to reduce the computing resources required for large-scale pre-training? 4) What aspect of knowledge do the pre-trained models provide for better language generation?

\par 
NLG tasks can be defined by the context features and mapping functions. The introduction of cross-lingual textual features \cite{Lample19} and task-specific Seq2Seq architectures \cite{SongTQLL19,Wang,Dong} in the pre-training stage has successfully boosted the performance on text-to-text generation. For NLG tasks concerning multiple modalities, it is conceivable that pre-training methods could also benefit from the joint consideration of cross-modal features. For example, in the vision-and-language field, the learning of cross-modal representations has proven to be highly effective \cite{LiuFenglin,LuJiasen}, but such representations can not yet be extracted from unpaired images and texts for image-grounded text generation, to the best of our knowledge. 

\par In NLU, it is possible to pre-train one model to obtain language representations once and for all. As for NLG, a task-agnostic pre-training algorithm should transcend the purpose of representation learning, and consider the general ability for language generation. The notion of ``encoder-agnostic adaption'' \cite{Ziegler19} makes a preliminary step towards this direction, but still remains far from approaching the equivalent performance as its NLU counterparts \cite{PetersNIGCLZ18,DevlinCLT19,Radford18,Yangzhilin}. 

\par Due to the colossal scale of the pre-training corpora, including a large number of parameters is essential to achieve favorable performance. As a result, the model size usually costs  at least 8 GPU cards
\cite{Dong,SongTQLL19,Lample19} in the pre-training for NLG systems, and it also hinders real-world applications. To reduce the memory consumption problem, existing work resorted to knowledge distillation to transfer the knowledge from a large teacher network to a small student network \cite{Sunsiqi,Jiaoxiaoqi}, or parameter reduction techniques to prune the model size in a more direct way \cite{LanZhenzhong}. However, the research context is limited to the NLU scenarios, and same endeavours are necessary to NLG applications. 

\par Another important branch of researches on unsupervised pre-training in NLP try to explain what kind of knowledge can be learned from pre-training. Related work has been done on the basis of both language understanding \cite{JawaharSS19,Petroni} and generation \cite{Abigail}. Specially, \cite{Abigail} analysed the characters of texts generated from a pre-trained GPT-2 by evaluating them over a wide spectrum of metrics. We argue that deeper understanding the way in which unsupervised pre-training contributes to better text generation, and the intrinsic mechanisms of the pre-trained models are also crucial to future work.
% We argue that understanding the information encoded in pre-trained models and its relationship with language generation is essential.

% some also investigate the inner functions that support the model to comprehension linguistic knowledge \cite{Betty19}. 
% In NLG, \cite{Abigail} studied the strength of the pre-trained models as text generator, by evaluating the characters of the generated stories.

\section{Conclusion}
\label{sec:conclusion}
Unsupervised pre-training has defined the state-of-the-arts on a variety NLP tasks. However, in the field of NLG, the diversity of context information is still impeding the the application of unsupervised pre-training. The major challenges exist in designing model architectures to cater for the assorted context, and retaining the general knowledge learned from pre-training. In this review, we survey the recent unsupervised methods to utilize large-scale corpora for NLG purposes, with a highlight on those aiming at facilitating the integration of pre-trained models with downstream tasks. We propose to classify them into architecture- and strategy-based methods, followed with detailed introductions and discussions of their pros and cons. Based on the comparison of these methods and analyses of some informative experimental results from previous publications, we summarize some scientific questions that has not yet been well understood, and suggest attention being paid to these questions by future work.

\bibliographystyle{abbrvnat}
\bibliography{neurips_2019}

\end{document}